\PassOptionsToPackage{table}{xcolor}
\documentclass[10pt,twocolumn,letterpaper]{article}
\usepackage{soul}
\usepackage{iccv}
\usepackage{times}
\usepackage{epsfig}
\usepackage{graphicx}
\usepackage{amsmath}
\usepackage{amssymb}
\usepackage{booktabs}
\usepackage{xspace}
\usepackage{caption}
\usepackage{graphicx} 
\usepackage{color}
\usepackage{algpseudocode}
\usepackage{algorithm}
\usepackage{threeparttable}
\usepackage[font=small,skip=1pt]{caption}
\usepackage{float}

\usepackage[pagebackref=true,breaklinks=true,letterpaper=true,colorlinks,bookmarks=false]{hyperref}

\iccvfinalcopy 

\usepackage{multirow} 
\usepackage{multicol}
\usepackage{graphicx}
\usepackage[export]{adjustbox}
\usepackage{tabularx}
\usepackage{colortbl}
\usepackage{hyperref}
\usepackage{rotating}
\usepackage{amsmath}
\usepackage[shortlabels]{enumitem}

\usepackage{tabularx}
\newcolumntype{C}[1]{>{\centering\let\newline\\\arraybackslash\hspace{0pt}}m{#1}} 
\newcolumntype{L}[1]{>{\let\newline\\\arraybackslash\hspace{0pt}}m{#1}} 
\usepackage[dvipsnames]{xcolor}
\ificcvfinal\fi

\begin{document}

\title{BOTT: Box Only Transformer Tracker for 3D Object Tracking}

\author{
Lubing~Zhou,~~Xiaoli Meng,~~Yiluan~Guo,~~Jiong Yang\\
Motional \\
{\tt\small \{lubing.zhou,~xiaoli.meng,~yiluan.guo,~jiong.yang\}@motional.com}
}

\maketitle
\ificcvfinal\thispagestyle{empty}\fi

\begin{abstract}
Tracking 3D objects is an important task in autonomous driving. Classical Kalman Filtering based methods are still the most popular solutions. However, these methods require handcrafted designs in motion modeling and can not benefit from the growing data amounts. In this paper, \textbf{B}ox \textbf{O}nly \textbf{T}ransformer \textbf{T}racker (BOTT) is proposed to learn to link 3D boxes of the same object from the different frames, by taking all the 3D boxes in a time window as input. Specifically, transformer self-attention is applied to exchange information between all the boxes to learn global-informative box embeddings. The similarity between these learned embeddings can be used to link the boxes of the same object. BOTT can be used for both online and offline tracking modes seamlessly. Its simplicity enables us to significantly reduce engineering efforts required by traditional Kalman Filtering based methods. Experiments show BOTT achieves competitive performance on two largest 3D MOT benchmarks: 
69.9 and 66.7 AMOTA on nuScenes validation and test splits, respectively, 56.45 and 59.57 MOTA L2 on Waymo Open Dataset validation and test splits, respectively. This work suggests that tracking 3D objects by learning features directly from 3D boxes using transformers is a simple yet effective way.
\end{abstract}
\section{Introduction}
\label{sec:introduction}
Autonomous driving is an open challenge attracting tremendous attention in the past decade. One of the most essential tasks for autonomous vehicles is to perceive 3D objects accurately, which includes the detection and tracking of the objects. Encouraging progress has been made in 3D object detection, owing to the emergence of large public multi-modality datasets~\cite{Caesar2020nuscenes,Sun2020WOD} and advanced 3D object detection methods ~\cite{bai2022transfusion,lang2019pointpillars,liu2022bevfusion,yin2021center}. On the other hand, tracking-by-detection methods for 3D Multi-Object Tracking (MOT)~\cite{kim2022polarmot,pang2021simpletrack,weng2020ab3dmot,zaech2022ogr3mot} remain competitive and popular due to their ability to benefit from powerful 3D object detectors. Among them, Kalman Filtering (KF) based trackers \cite{pang2021simpletrack,weng2020ab3dmot} are dominant, as the kinematics models are naturally designed for tracking 3D motion.

Despite their competitiveness, KF-based trackers have two main disadvantages. Firstly, a series of Kalman filters must be defined to cover various types of motion kinematics, including static, constant velocity, constant acceleration, constant angular velocity, and more sophisticated non-constant ones. Meanwhile, Kalman filters require specific parameters for each object category, such as the mean and variance of measurements and noises. 
Hence, KF-based trackers need much engineering efforts to tune these parameters to have a decent performance. Second, KF-based trackers could not make use of modern large datasets~\cite{Caesar2020nuscenes,Sun2020WOD} to boost the performance.

One approach for data-driven 3D MOT is to perform joint detection and tracking in a single stage, such as SimTrack~\cite{luo2021exploring} and CenterPoint~\cite{yin2021center}. While these methods can simultaneously detect and track 3D objects in point clouds with a single model, the tracking is normally limited to consecutive frames to fit the architecture of the 3D object detector based on lidar. However, there is a fundamental conflict between the two tasks: 3D detection focuses on the instantaneous spatial localization of objects with only a few point clouds, while 3D tracking requires a much longer spatial-temporal memory. In practice, 4D spatial and temporal learning with significantly more point clouds is still an open challenge due to computational complexity and hardware limitation.

\begin{figure*}[!ht]
\begin{center}
\includegraphics[width=0.8\linewidth]{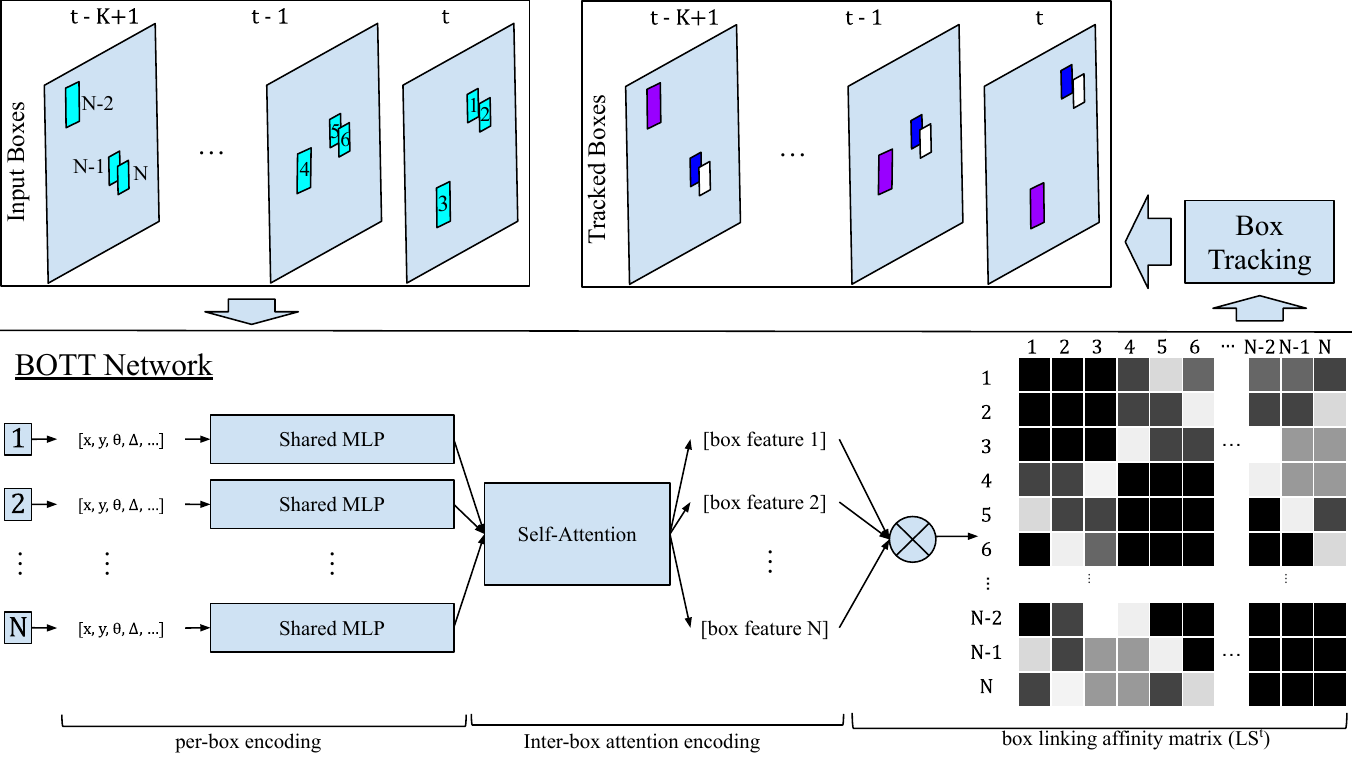}
\end{center}
\caption{BOTT 3D tracking architecture. It contains the BOTT network and the box tracking module. BOTT network consumes all boxes from $K$ frames to generate a pairwise box affinity matrix with 3 steps: per-box encoding, inter-box attention encoding, and affinity matrix generation by dot-product. The box tracking module produces tracks based on the pairwise box linking scores to tracks. Both online and offline tracking are supported.}
\label{fig:bott_overview}
\end{figure*}

An alternative research direction is to learn to track the bounding boxes of the 3D objects directly~\cite{hung2020soda,kim2022polarmot,zaech2022ogr3mot}. This approach offers a straightforward replacement of KF-based trackers in the existing tracking-by-detection paradigm. Machine learning methods, consuming only the geometric properties of bounding boxes, inherit the merits of KF-based trackers and could benefit from growing data amounts. However, 3D box-based learning methods face two key challenges. Firstly, each input frame contains a varying number of unordered boxes, making it difficult to establish a consistent identity for each object. Secondly, unlike image appearance features, 3D box geometric features lack spatial-temporal consistency for each object identity. 
Nevertheless, humans can easily connect boxes from the same object when viewing the bird's-eye view boxes sequentially by interpreting the global box distributions and the spatial-temporal context of each individual box. 
In other words, the box features, i.e. position, size, orientation, object types, and their temporal-spatial distributions should be sufficient for tracking. The key is to find a suitable tool to learn such information for each box. PolarMOT~\cite{kim2022polarmot} is an inspiring work in this regard, which uses a graph neural network (GNN) to iteratively learn box features from spatial-temporal local boxes. Differently, we propose a novel approach, called the \textbf{B}ox \textbf{O}nly \textbf{T}ransformer \textbf{T}racker (BOTT), that uses attention~\cite{vaswani2017attention} to globally learn per-box embeddings from all multi-class boxes with a single model, as shown in Figure.~\ref{fig:bott_overview}. BOTT is well-suited for 3D box tracking, as attention mechanisms have repeatedly demonstrated its effectiveness in communicating information temporally and spatially between inputs of varying-length~\cite{bai2022transfusion,meinhardt2022trackformer,zhou2022global}. 

In summary, the proposed BOTT is a simple multi-class multi-object 3D box-only tracker. Attentive box features are globally learned to encode box information and its spatial-temporal distributions among other boxes. The linking scores or similarity scores between box features are used to link boxes. Our main contributions are as follows:
\begin{itemize}[topsep=0pt]
    \item We propose BOTT, a simple self-attention based tracker, which consumes only 3D bounding boxes. The simplicity and effectiveness pave the path for more potential works to track 3D boxes using transformers. Meanwhile, it could be extended to other applications such as multi-modal 3D box tracking. 
    \item We provide complete online and offline tracking algorithms for multi-class 3D tracking with a single model under the BOTT framework. 
    \item We conduct experiments on two largest 3D MOT datasets nuScenes\cite{Caesar2020nuscenes} and Waymo Open Dataset (WOD) \cite{Sun2020WOD}. BOTT demonstrates competitive performance: 69.9 and 66.7 AMOTA for nuScenes val and test splits, 56.45 and 59.57 MOTA L2 for WOD val and test splits.
    \item We conduct extensive ablation studies to examine the key designs that enable strong performance and validate the BOTT's generalization ability across datasets and input frequencies.
\end{itemize}
\section{Related Work}
\label{sec:related_work}

This section first reviews the 3D MOT algorithms based on the tracking-by-detection paradigm, then reviews the transformer-based trackers, and lastly reviews online and offline MOT.  

\subsection{3D MOT}
Under the tracking-by-detection framework, AB3DMOT~\cite{weng2020ab3dmot} serves as a baseline, using a simple KF tracking framework. Many methods have been proposed to improve the tracking performance based on the same KF-based tracking framework, such as ProbTrack~\cite{Chiu2022prob3d} and SimpleTrack~\cite{pang2021simpletrack}. Their major difference lies in the association metrics: AB3DMOT used 3D Intersection of Union (IoU); ProbTrack used Mahalabnobis distance; and SimpleTrack used 3D generalized 3D (GIoU) while a second association was applied for better handling lower-score objects.

Lately, more learning-based tracking algorithms have been reported. Some of them proposed to use GNNs~\cite{buechner2022batch3dmot,kim2022polarmot,weng2020gnn3dmot, zaech2022ogr3mot} as graphs are a natural representation of the MOT problem, where the detected objects are encoded as nodes and their spatial-temporal relations are represented by edges. GNN3DMOT~\cite{weng2020gnn3dmot} used GNNs to estimate an affinity matrix and solved the association using the Hungarian algorithm. OGR3MOT~\cite{zaech2022ogr3mot} solved 3D MOT with GNNs in an end-to-end manner, focusing on the data association and the classification of active tracklets. Batch3dmot~\cite{buechner2022batch3dmot} presented a novel multi-modal GNN framework for offline 3D MOT on multi-class tracking graphs that introduces $k$ nearest neighborhood attention across graph components to allow information exchange across disconnected graph components. Closest to our approach, PolarMOT~\cite{kim2022polarmot} explored the impact of geometric relationships between objects based on GNNs with solely 3D boxes as well. The difference is that in the PolarMOT, mutual object interactions within local regions are learned implicitly through iterative message-passing steps, while we used self-attention to get the global context information in one shot. 

\subsection{Transformer Tracker}
Over the past few years, transformer networks have gained great momentum for their ability to effectively process sequence data. Encouraging tracking performance on 2D MOT has been demonstrated from transformer trackers with image appearance features~\cite{hung2020soda,meinhardt2022trackformer,ruppel2022transmot,sun2020transtrack,zhou2022global,zhu2021mo3tr}. This is mainly because transformers could handle long-term dependencies and is robust to occlusions and complex scenarios. Among these trackers, SODA~\cite{hung2020soda} and GTR~\cite{zhou2022global} are more related to our work. 
SODA proposed an attention measurement encoding to compute track embeddings from objects' appearance features to reason about the spatial-temporal dependencies between all objects. GTR fed the objects' appearance features into the encoder of the transformer, additionally took trajectory queries as the decoder input, and produced association scores between each query and object. Different from their work, our work learns the spatial-temporal context information from 3D bounding boxes only without appearance using simple self-attention encoders.

\subsection{Offline and Online Tracking}
Recently, offline auto-labeling methods~\cite{Yang2021auto4d,qi2021offboard} have drawn great attention in the autonomous driving industry, as they could scale up the data annotation drastically. In the auto-labeling pipeline, future information could be introduced to improve tracking performance. It is not straightforward for KF-based trackers~\cite{Chiu2022prob3d, pang2021simpletrack, weng2020ab3dmot} to leverage on the future information, as they are designed to operate recursively and rely solely on the current state and observations. In contrast, BOTT provides a convenient solution for both online and offline multi-class tracking. 

\section{Box Only Transformer Tracker}
In this section, the proposed BOTT framework will be presented, including the BOTT network and box tracking algorithms. 
\label{sec:bott}
\subsection{Overview}
In a scene with $T$ frames $\{I^1, I^2, \cdots, I^T\}$, there are $n_t$ detected 3D boxes for frame $I^t$: $B^t \mathrm{=} \{ b^t_{1}, b^t_{2}, \cdots, b^t_{n_t} \}$. Each box $b^t_{i}$ contains raw features including $(x^t_i, y^t_i, z^t_i, w^t_i, l^t_i, h^t_i, \theta^t_i, t, c^t_{1i}, c^t_{2i}, \cdots, c^t_{Ci})$ ($xyz$: center, $wlh$: size, $\theta$: yaw, $t$: time, $c_1{\cdots}c_C$: classification scores for $C$ categories). A sliding window $S^t_{K}$ is defined as a set of all the boxes from $K$ consecutive frames: $S^t_{K} \mathrm{=} \{B^{t-K+1}, \cdots, B^{t-1}, B^t\}$. For simplicity, $S^{t}$ denotes a sliding window with latest frame $I^t$ without explicit statement of $K$, and $N_{t}\mathrm{=}\sum_{j=0:K-1}n_{t-j}$ denotes the total box number within $S^t$. 

As shown in Figure.~\ref{fig:bott_overview}, BOTT accepts sliding windows as input and estimates pairwise box linking scores using a BOTT network. Then a box tracking module connects the boxes to produce tracks using the linking scores. The BOTT network contains three components: 1) single-box feature encoding; 2) inter-box encoding with self-attention; and 3) linking scores estimation.

\subsection{Single-Box Feature Encoding}
\label{sec:box_encod}
This module aims to learn high-level per-box features from raw geometric features. The raw center values $xyz$  may vary greatly in different sliding windows because they are in the shared global coordinates. To reduce variance, center positions of all boxes in $S^t$ are normalized by subtracting the minimal center values $(x^{t}_{min}, y^{t}_{min}, z^{t}_{min})$ of all boxes. Take $x^{t}_{min}$ for instance, $x^{t}_{min} = \min_{\{(t, i)|b^t_{i} \in S^t\}}{x^t_i}$. The time feature $\Delta_i^t$ of box $b_i^t$ is encoded as the delta between the box frame and center frame in $S_t$. Box heading angle is encoded as $[\sin(\theta^t_i), \cos(\theta^t_i)]$. In summary, the input feature of $b^t_{i}$ includes $(x^t_i - x^{t}_{min}, y^t_i-y^{t}_{min}, z^t_i-z^{t}_{min}, w^t_i, l^t_i, h^t_i, \sin(\theta^t_i), \cos(\theta^t_i), \Delta_i^t, c^t_{1i}, c^t_{2i}, \cdots, c^t_{Ci})$. Each box feature vector is fed into a shared Multi-Layer Perception (MLP), which has $N_{MLP}$ latent layers with $d_l$ dimensions and a final output layer with $d$ dimensions. $N_t\times d$ feature embeddings are learned for all boxes $S^t$.
 
\subsection{Inter-Box Encoding with Self-Attention}
After per-box feature encoding, the $N_t \times d$ output embeddings are fed to a self-attention module to encode inter-box relationship. Specifically, $N_{enc}$ identical transformer encoder blocks~\cite{vaswani2017attention} are applied sequentially to exchange information between all input box embeddings. Each encoder block consists of a multi-head attention network with $n_h$ heads and a following feed-forward network with $d_{f}$ hidden nodes. After self-attention, the size of output box embeddings remains $N_t\times d$ for $S^t$.

Notably, the self-attention in BOTT is class-agnostic, i.e. each box learns to get information from all the other boxes in the sliding window. 
As shown in Figure.~\ref{fig:attentions}, the moving car in sub-figure (a) and (b) also gets attention from nearby pedestrians, and the pedestrian in sub-figure (c) and (d) also gets attention from nearby cars. Self-attention is expected to learn robust box representation by encoding global spatial-temporal box distributions into each box. Another advantage of class-agnostic self-attention is that it enables BOTT to handle multi-class objects tracking with a single model. It greatly reduces the deployment complexity in practice.

\subsection{Linking Score Estimation}
Tracked boxes from the same object share a similar spatial-temporal context within a sliding window. With the learned $N_t\times d$ box embeddings, $L_2$ normalization is performed to normalize each box embedding to unit length. Then, simple dot-product is used to compute inter-box linking scores, resulting in a $N_t\times N_t$ linking score matrix $LS^t$ (illustrated in Figure.~\ref{fig:bott_overview}). For convention, the linking score is normalized to range $[0, 1]$ by $LS^t \mathrm{=} \frac{LS^t + 1}{2}$. A linking score of $1$ indicates that two boxes belong to the same object, and $0$ means the opposite. In this way, the tracking is converted to a binary classification problem of box links.

\subsection{Loss}
\label{sec:loss}
Assume $y^t \in \mathcal{R}^{N_t{\times}N_t}$ denote the corresponding ground truth linking score matrix for $LS^t$, binary cross entropy between them is used as the loss. During training, to prevent the easy cases from overwhelming the loss, a binary mask $M^t\in\mathcal{R}^{N_t{\times}N_t}$ is constructed to ignore losses from: 1) inter-class box links; 2) box links from the same frame; 3) box links between two false positive boxes (detected boxes without associated ground truth boxes); and 4) box links whose center distances exceed the expected maximum displacement for the object over the given time duration. 

Given a batch data of $b$ sliding windows $\{S^t, t \in \{t_1, t_2,\cdots, t_b\}\}$, the linking loss $L_{link}$ is computed as:
\begin{align}
    L^t_{ij} & = (1-\beta)(1 - y^t_{ij})\log(1 - LS_{ij}^t)  + \beta y^t_{ij} \log LS_{ij}^t \nonumber  \\
    L_{link} & = \frac{\sum_{t} \sum_{i,j} (M^t_{ij} \cdot L^t_{ij})}{\sum_t \sum_{i,j} M^t_{ij}} 
\label{eq:loss}
\end{align}
where $\beta$ is the positive sample weight.

\subsection{Box Tracking with BOTT}
Linking scores are utilized to create tracks in box tracking module. Depending on whether future data can be used, BOTT can perform both online and offline tracking.

\subsubsection{Online Tracking}
\label{sec:online_tracker}
\begin{figure*}[!htb]
\begin{center}
\includegraphics[width=0.9\linewidth]{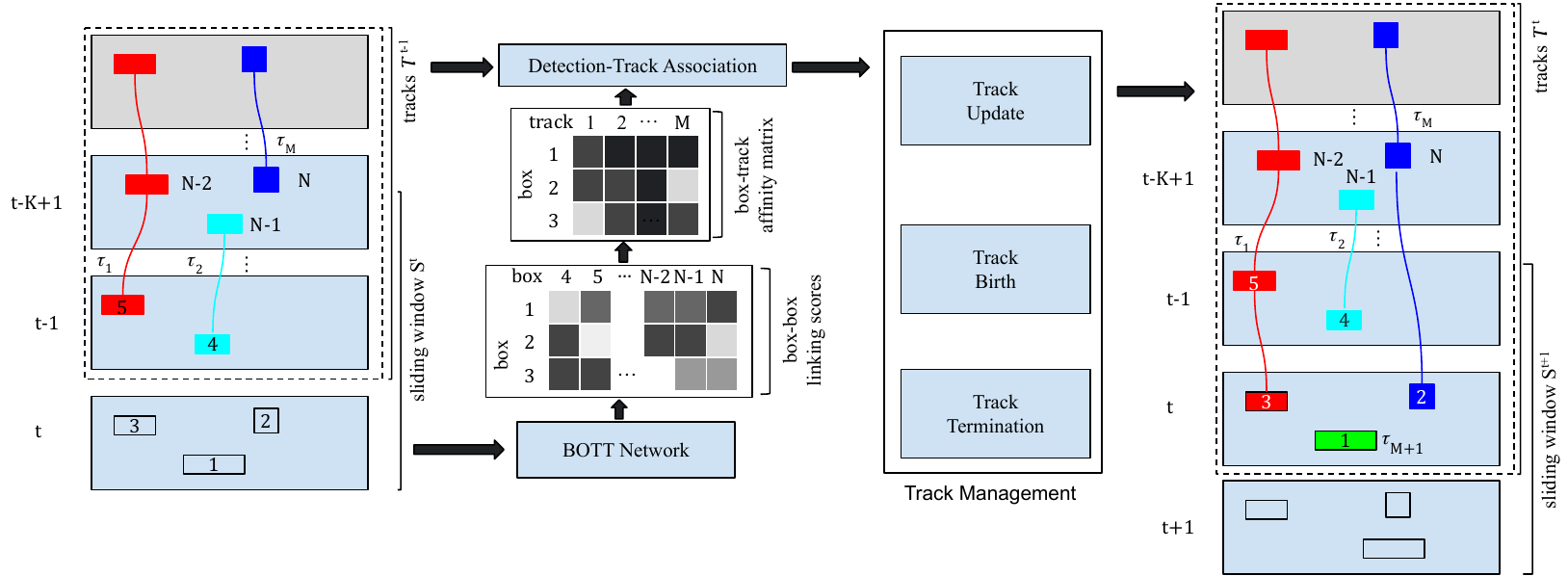}
\end{center}
   \caption{BOTT online tracking architecture. BOTT network estimates the linking scores for latest sliding window. Box-to-box scores are converted to box-to-track affinity score for detection-track association. Track birth, update and termination are managed by a track management module.}
\label{fig:bott_online_tracking}
\end{figure*}

Figure.~\ref{fig:bott_online_tracking} illustrates online box tracking under BOTT framework. At time $t$, the set of history tracks are denoted as $\mathcal{T}^{t-1}\mathrm{=}\{\tau_1, \tau_2, \cdots, \tau_M\}$, where a track $\tau_i$ is defined as a list of time-ordered 3D boxes. In the example of Figure.~\ref{fig:bott_online_tracking}, $\tau_1\mathrm{=}[\cdots,b_{N-2},\cdots,b_{5}], \tau_2\mathrm{=}[b_{N-1},\cdots, b_{4}]$, where $b_i$ denotes the $i^{th}$ box as marked. At frame $I^t$, the latest sliding window $S^t_K$ is fed into BOTT network to generate linking scores $LS^t$ between all the boxes in $S^t_K$, i.e. $\{B^{t-K+1}, \cdots, B^{t-1}, B^t$\}. Since the goal of online tracking is to connect new detections $B^t$ to existing tracks $\mathcal{T}^{t-1}$, we only use linking scores between boxes $B^t$ (i.e., $\{b_1, b_2, b_3\}$) and boxes $\{B^{t-K+1}, \cdots, B^{t-1}\}$ (i.e., $\{b_4, b_5,\cdots,b_N\}$). Moreover, linking scores are set to zero for box links under conditions: 1) the two boxes are from different object categories; 2) the center distance violates the physical constraints as discussed in section~\ref{sec:loss}. Next, the affinity score $AS_{ij}$ between each detection $b_i \in B^t$ and each track $\tau_j\in \mathcal{T}^{t-1}$ is calculated as maximum of all box pairs:
\begin{equation}
    AS^{t}_{ij} = \max_{k\in\{k|b_k \in ( \tau_j \cap S^t)\}} LS^t_{ik}
\end{equation}
Instead of track propagation by motion model and association in latest frame, BOTT directly conducts Hungarian association between all detections and tracks using cost matrix $1-AS^t$. This enables BOTT to handle multi-class tracking with single model.

BOTT online tracking utilizes a simple track management strategy to control birth, update and termination of tracks. Each associated detection is appended as the new tail box in associated track, sharing the same track identity. Unmatched detections will birth a track with \textit{unconfirmed} status. An unconfirmed track will be changed to \textit{confirmed} status after it has accumulated minimal $N_{birth}$ boxes, and a confirmed track will be terminated after $T_{term}$ seconds without new detections. This work uses $N_{birth}\mathrm{=}1$ and $T_{term} \mathrm{=}2$. After track management, tail boxes of confirmed tracks in frame $I^t$ will be published as the latest tracking result.

\subsubsection{Offline Tracking}
BOTT can also be used to perform offline tracking. Thanks to the effectiveness of BOTT, a simple greedy approach is enough to achieve promising performance. In offline setting, all sliding windows $S^t$ are first constructed with a stride of one, and fed to the BOTT network to generate linking scores. Let $LS^t(b_i, b_j)$ denote the estimated linking score between box $b_i$ and $b_j$ in sliding window $S^t$. The linking score between $b_i$ and $b_j$ across the scene is computed as:
\begin{equation}
  \hat{LS}(b_i, b_j) = \max_{t\in[1,T], (b_i, b_j) \in S^t} LS^t(b_i, b_j)  
\end{equation}
 And an optimal threshold is applied to remove links with low $\hat{LS}$. To remove redundant links, non-maximum suppression is performed according to $\hat{LS}$. For example, if a link between $b_i$ and $b_j$ from two frames has been selected, all the other links to $b_i$ or $b_j$ between the same two frames will be pruned. To further prevent false links, boxes that violate physical constrains as in online tracking or boxes from different classes are pruned. Finally, box interpolations are also performed to fill any gaps between the linked boxes.
\section{Experimental Setup}
\label{sec:exp_setup}
\subsection{Datasets and Metrics}
We evaluate BOTT on the two largest benchmarks for 3D MOT: nuScenes~\cite{Caesar2020nuscenes} and Waymo Open Dataset (WOD)~\cite{Sun2020WOD}. 
\paragraph{nuScenes} consists of 1000 driving scenes of approximately 20 seconds long. $700/150/150$ scenes are used for training, validation, and test, respectively. Lidar scans and ground truth (GT) 3D box annotations are provided at 20Hz and 2Hz, respectively. We report the overall Average Multiple Object Tracking Accuracy (AMOTA)~\cite{weng2020ab3dmot}, recall, and identity switches(IDS) across all the 7 tracking categories, i.e. car, pedestrian, bicycle, bus, motorcycle, trailer, and truck, and also the AMOTA for each individual category. Overall AMOTA is the primary metric for nuScenes benchmark.
\paragraph{Waymo Open Dataset} contains 1000 driving sequences of 20 seconds, with $798/202/150$ sequences for training, validation and test, respectively. Both point clouds and GT 3D boxes of objects are provided at 10Hz. We report MOTA for both the L1 and L2 difficulty levels, mismatch ratio~\cite{Sun2020WOD} for objects in the L2 difficulty, and MOTA for all the 3 tracking categories (vehicle, pedestrian and cyclist) in the L2 difficulty level. MOTA in the L2 difficulty level is the primary metric for WOD 3D MOT benchmark.

\subsection{Track Database Generation}
\label{sec:track_db_gen}
CenterPoint~\cite{yin2021center} is deployed on the training, validation and test sets on both nuScenes and WOD to get the detections to generate a track database. 
With the detections, a Non-Maximum Suppression (NMS) is applied to remove overlapped boxes, and boxes with detection scores below a threshold are also filtered. 
To generate the database, an association between detections and GT boxes provided by nuScenes or WOD is performed. The track IDs in the GT boxes are assigned to the associated detection boxes, and unmatched detection boxes are considered as false positives. 
For both nuScenes and WOD, the track database are generated at 10Hz. The 2Hz GT provided by nuScenes is interpolated to get 10Hz GT. Each scene in the track database is divided into overlapping $K-$frame sliding windows with a stride of one. This work uses $K\mathrm{=}{16}$. 
\subsection{Implementation Details}
\subsubsection{Network Details}
The MLP in single box encoding consists of four Linear$+$ReLU blocks, with output dimensions $(1024, 1024, 1024, 512)$. Three 
stacked identical encoder blocks are used for inter-box encoding, with each block including 1) a $8-$head attention
following a LayerNorm~\cite{ba2016layernorm}; and 2) a feature feed-forward net with two Linear$+$ReLU blocks (output dimensions $(1024, 512)$), following a LayerNorm. 
The output box embeddings remains the size of $N_t\times d=512$. 
\subsubsection{Training Procedure}
The link distribution is very imbalanced in sliding windows, with over $90\%$ negative links. Hard negative sample mining is applied to pick all positives (assume $P$ links) and maximally $\kappa P$ negative links with largest linking score errors ($\kappa\mathrm{=}{4}$ is used). 
The positive weight $\beta$ in Eq.(\ref{eq:loss}) is set to $\frac{\kappa}{\kappa+1}$.
BOTT is trained with Adam optimizer~\cite{kingma2014adam} for 50 epochs with batch size of 4 sliding windows. We use the 1cycle learning rate policy~\cite{smith20171cyclr} with intial learning rate of $1\mathrm{e}{-3}$. As each sliding window has varying number of boxes, we apply zero padding to each sliding window to have a largest box number of all sliding windows in a batch, and a binary masking is used to prohibit attention from padded boxes in attention layers.
\subsubsection{Data Augmentation}
First, we drop some tracks randomly in the sliding windows to reduce the number of boxes to a maximum number 3000, to mimic occlusions and false negatives. Next, all boxes in a sliding window will be shifted from global coordinates to local coordinates centered at the middle of all box centers, i.e. $ [(\text{max}_x + \text{min}_x) /2, (\text{max}_y + \text{min}_y)/2) ]$. Finally, we perform two set of global augmentations to all the boxes in a sliding window: we first apply a random flip along x axis and$/$or y axis, then a global rotation (yaw uniformly drawn from $[-\pi/2, \pi/2]$).

\subsection{Public Benchmark}

Tracking performance heavily relies on the detection quality. For fair comparison, we compare BOTT with these published trackers that are also based on the commonly used CenterPoint detections~\cite{yin2021center}. Among them, AB3DMOT~\cite{weng2020ab3dmot} and SimpleTrack~\cite{pang2021simpletrack} trackers belongs to classic motion trackers, while PolarMOT~\cite{kim2022polarmot}, OGR3MOT~\cite{zaech2022ogr3mot}, CenterPoint~\cite{yin2021center} and Batch3DMOT~\cite{buechner2022batch3dmot} are learning based trackers.
Table.~\ref{tab:nuscenes_benchmark} shows the 3D MOT results on nuScenes validation and test sets. In the validation set, we compare the performance for both online and offline tracking. Our online BOTT achieves a 2.53 AMOTA improvement over learning based tracker PolarMOT, and achieves slightly better performance that the advanced classic motion trackers, i.e. SimpleTrack~\cite{pang2021simpletrack}. Our offline BOTT achieves state-of-the-art performance of the Lidar box-based offline tracking algorithms, with 71.38 AMOTA. One should note that the tracking performance for Batch3DMOT~\cite{buechner2022batch3dmot} is presented with box feature only. For the test set, only online tracking results are compared between different trackers. Similarly, our online BOTT achieves better performance than learning based box tracker, and comparable results with classic motion trackers.

Table.~\ref{tab:waymo_benchmark} shows the 3D MOT results on WOD. Our online BOTT could achieve better results than learning based trackers, i.e. CenterPoint~\cite{yin2021center}, and achieves comparable performance to advanced classic motion trackers, i.e. SimpleTrack~\cite{pang2021simpletrack}.
\begin{table*}[!htb]
\centering
\footnotesize
\begin{threeparttable}
\caption{nuScenes 3D MOT benchmark results. Best result is marked in \textbf{bold}, and best ML based result is shaded.}
\label{tab:nuscenes_benchmark}
\setlength{\tabcolsep}{3pt}
\begin{tabular}
{|l|l|l|ccc|ccccccc|}
\hline
& \multirow{2}{*}{Method} & \multirow{2}{*}{Modality} & \multirow{2}{*}{AMOTA$\uparrow$} & \multirow{2}{*}{IDS$\downarrow$} & \multirow{2}{*}{Recall~$\uparrow$} & \multicolumn{7}{c|}{class-specific AMOTA$\uparrow$} \\
& & &  &  &  & car & ped & bicycle & bus & motor & trailer & truck \\
\hline\hline
\multirow{8}{*}{\rotatebox[origin=c]{90}{val set}} & AB3DMOT\textsuperscript{*}~\cite{weng2020ab3dmot} & Box3D  & 57.8 & 1275 & - 
& - & - & - & - & - & - & - \\ [-0.06cm]
& ProbTrack\textsuperscript{*}\cite{Chiu2022prob3d}& Box3D & 62.4 & 1098 & - 
 & 73.5 & 75.5 & 27.2 & 74.1 & 50.6 & 33.7 & 58.0 \\ [-0.06cm]
& SimpleTrack\textsuperscript{*}~\cite{pang2021simpletrack} & Box3D  & 69.57 & \textbf{403} & \textbf{73.61} 
& \textbf{83.9} & \textbf{80.67} & 50.3 & 79.52 & \textbf{74.19} & \textbf{52.7} & 65.69 \\[0.12cm]
& CenterPoint~\cite{yin2021center}& Lidar & 66.75 & 616 & 70.5 
 & \colorbox{gray!25}{83.83} & 77.02 & 47.56 & \colorbox{gray!15}{\textbf{84.83}} & 60.22 & 45.73 & 68.02 \\ [-0.08cm]
& Online PolarMOT~\cite{kim2022polarmot} & Box3D & 67.27 & 439 & \colorbox{gray!25}{72.46} 
& 81.26 & 78.79 & 49.38 & 82.76 & 67.19 & 45.8 & 65.7\\ [-0.08cm]
& Online BOTT & Box3D & \colorbox{gray!15}{\textbf{69.91}} & \colorbox{gray!25}{438} & 72.08 & 83.67 & \colorbox{gray!25}{80.02} & \colorbox{gray!15}{\textbf{50.34}} & 83.66 & \colorbox{gray!25}{71.53} & \colorbox{gray!25}{51.34} & \colorbox{gray!15}{\textbf{68.85}} \\ [0.05cm]
\cline{2-13}
& Offline Batch3DMOT~\cite{buechner2022batch3dmot} & Box3D & 70.6 & 758 & 72.0 
& 83.4 & 81.1 & \colorbox{gray!15}{\textbf{54.8}} & 83.5 & 73.3 & 49.6 & 68.5\\ [-0.06cm]
& Offline PolarMOT~\cite{kim2022polarmot} & Box3D & 71.14 & \colorbox{gray!15}{\textbf{213}} & \colorbox{gray!15}{\textbf{75.14} } & \colorbox{gray!15}{\textbf{85.83}} & 81.7 & 54.1 & \colorbox{gray!15}{\textbf{87.36}} & 72.32 & 48.67 & 68.03\\ [-0.06cm]
& Offline BOTT & Box3D & \colorbox{gray!15}{\textbf{71.38}} & 310 & 73.3 & 84.3 & \colorbox{gray!15}{\textbf{82.1}} & 53.8 & 85.4 & \colorbox{gray!15}{\textbf{74.2}} & \colorbox{gray!15}{\textbf{51.2}} & \colorbox{gray!15}{\textbf{68.7}} \\ [0.05cm]
\hline
\multirow{4}{*}{\rotatebox[origin=c]{90}{test set}} & AB3DMOT\textsuperscript{*}~\cite{weng2020ab3dmot} & Box3D & 15.1 & 9027 & 27.6 & 27.8 & 14.1 & 0 & 40.8 & 8.1 & 13.6 & 1.3 \\ [-0.06cm]
& ProbTrack\textsuperscript{*}~\cite{Chiu2022prob3d} & Box3D & 55.0 & 950 & 76.8 & 71.9 & 74.5 & 25.5 & 64.1 & 48.1 & 49.5 & 51.3 \\ [-0.06cm]
& SimpleTrack\textsuperscript{*}~\cite{pang2021simpletrack} & Box3D & \textbf{66.8} & 575 & \textbf{70.3} & 82.3 & 79.6 & \textbf{40.7} & 71.5 & 67.4 & 67.3 & 58.7 \\ [0.12cm]
& CenterPoint~\cite{yin2021center} & Lidar &  63.8 & 760 & 67.5 & 82.9 & 76.7 & 32.1 & 71.1 & 59.1 & 65.1 & 59.9 \\ [-0.06cm]
& PolarMOT~\cite{kim2022polarmot} & Box3D & 66.4 & \colorbox{gray!15}{\textbf{242}} & \colorbox{gray!15}{\textbf{70.2}} & \colorbox{gray!15}{\textbf{85.3}} & \colorbox{gray!15}{\textbf{80.6}} & 34.9 & 70.8 & \colorbox{gray!15}{\textbf{65.6}} & 67.3 & 60.2 \\ [-0.06cm]
& OGR3MOT~\cite{zaech2022ogr3mot} & Box3D & 65.6 & 288 & 69.2 & 81.6 & 78.7 & \colorbox{gray!25}{38.0} & 71.1 & 64.0 & 67.1 & 59.0  \\ [-0.06cm]
& Online BOTT & Box3D & \colorbox{gray!25}{66.7} & 743 & 67.7 & 83.1 & 75.9 & 32.6 & \colorbox{gray!15}{\textbf{74.8}} & 65.8 & \colorbox{gray!15}{\textbf{70.1}} & \colorbox{gray!15}{\textbf{64.3}} \\
\hline
\end{tabular}
\begin{tablenotes}
\item *Non-ML methods
\end{tablenotes}
\vspace{-0.3cm}
\end{threeparttable}
\end{table*}
\section{Results}
\label{sec:results}

\begin{table*}[!htb]
\centering
\footnotesize
\begin{threeparttable}
\caption{WOD 3D MOT benchmark results. Best result is marked in \textbf{bold}, and best ML based result is shaded.}
\label{tab:waymo_benchmark}
\setlength{\tabcolsep}{3pt}
\begin{tabular}
{|l|l|l|ccc|ccc|}
\hline
& \multirow{2}{*}{Method} & \multirow{2}{*}{Modality} & \multirow{2}{*}{MOTA(L1)$\uparrow$} & \multirow{2}{*}{MOTA(L2)$\uparrow$}  & \multirow{2}{*}{Mismatch~$\downarrow$} & \multicolumn{3}{c|}{class-specific MOTA(L2)$\uparrow$} \\
& & &  &  &  & vehicle & pedestrian & cyclist \\
\hline\hline
\multirow{6}{*}{\rotatebox[origin=c]{90}{val set}} & AB3DMOT\textsuperscript{*}~\cite{weng2020ab3dmot} & Box3D & - & - & - 
& 40.1 & 37.7 & -\\ [-0.06cm]
& ProbTrack\textsuperscript{*}~\cite{Chiu2022prob3d} & Box3D & - & - & - 
& 54.06 & 48.10 & - \\ [-0.06cm]
& SimpleTrack\textsuperscript{*}~\cite{pang2021simpletrack}& Box3D & \textbf{59.44} & \textbf{56.92} & 0.36 & \textbf{56.12} & \textbf{57.76} & 56.88 \\[0.1cm]
& CenterPoint~\cite{yin2021center}& Lidar & 58.35 & 55.81 & 0.74
 & 55.05 & 54.94 & 57.44\\ [-0.06cm]
& Online BOTT & Box3D & \colorbox{gray!25}{58.98} & \colorbox{gray!25}{56.50} & \textbf{0.35} & \colorbox{gray!25}{55.11} & \colorbox{gray!25}{56.48} & \colorbox{gray!15}{\textbf{57.78}} \\
\cline{2-9}
& Offline BOTT & Box3D & 59.67 & 57.14 & 0.12 & 55.17 & 57.05 & 59.21 \\
\hline
\multirow{5}{*}{\rotatebox[origin=c]{90}{test set}} & AB3DMOT\textsuperscript{*}~\cite{weng2020ab3dmot} & Box3D & - & - & - & 57.73 & 53.80  & - \\ [-0.06cm]
& ProbTrack\textsuperscript{*}~\cite{Chiu2022prob3d} & Box3D & 49.16 & 47.65 & 1.01 
& 49.32 & 44.38 & 25.29\\ [-0.06cm]
& SimpleTrack\textsuperscript{*}~\cite{pang2021simpletrack} & Box3D & \textbf{61.82} & \textbf{60.18} & 0.38 & \textbf{60.3} & \textbf{60.13} & 60.12 \\[0.1cm]
& CenterPoint~\cite{yin2021center} & Lidar & 60.31 & 58.67 & 0.72 & 59.38 & 56.64 & 60.0 \\ [-0.06cm]
& Online BOTT & Box3D & \colorbox{gray!25}{61.20} & \colorbox{gray!25}{59.57} & \colorbox{gray!25}{\textbf{0.31}} & \colorbox{gray!25}{59.49} & \colorbox{gray!25}{58.82} &  \colorbox{gray!15}{\textbf{60.41}} \\
\hline
\end{tabular}
\begin{tablenotes}
\item *Non-ML methods
\end{tablenotes}
\end{threeparttable}
\vspace{-0.2cm}
\end{table*}

\subsection{Qualitative Analysis}
\begin{figure}[!htb]
\begin{center}
\includegraphics[width=\linewidth]{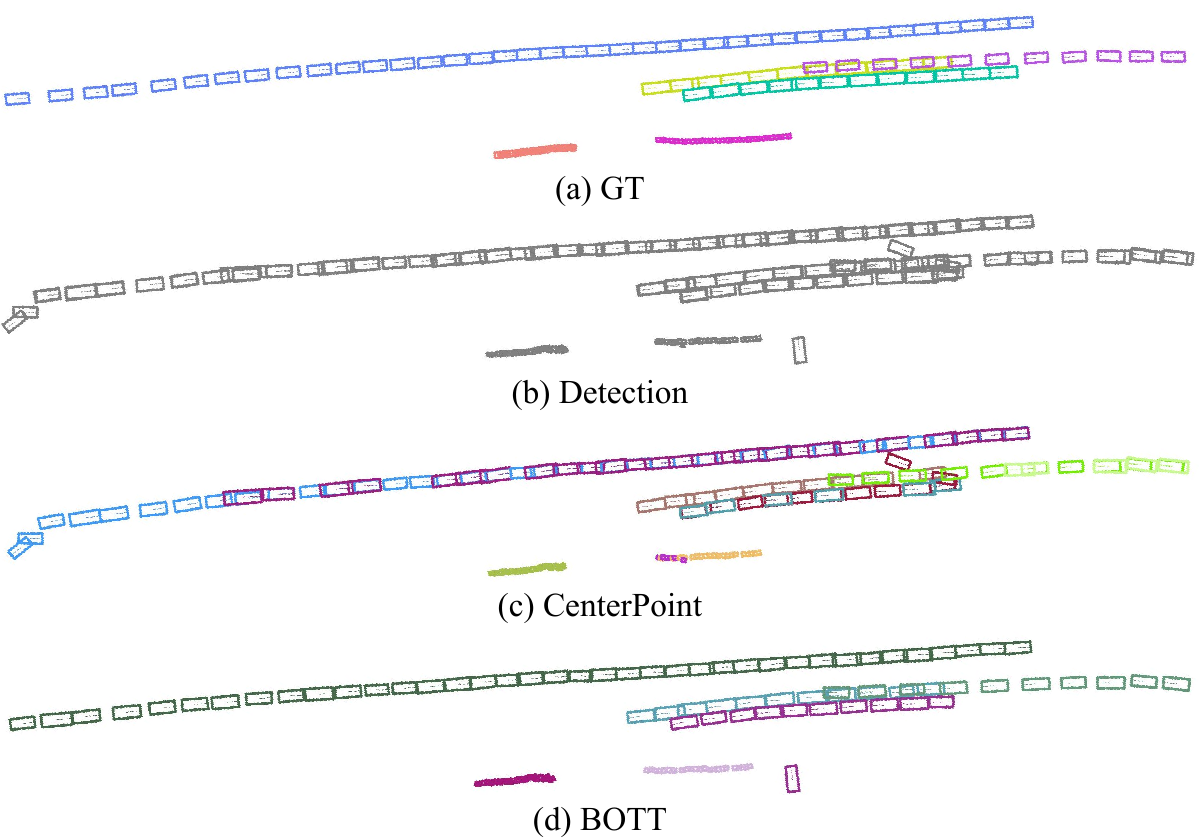}
\end{center}
   \caption{Example of BOTT tracking results.
   }
\label{fig:bott_better}
\vspace{-0.2cm}
\end{figure}

Figure.~\ref{fig:bott_better} illustrates the qualitative results of BOTT which accumulate over 40 frames in a scene on the nuScenes validation set. Figure.~\ref{fig:bott_better} (a) and (b) shows GT and raw detections from CenterPoint~\cite{yin2021center}, respectively, and (c) and (d) shows the tracking results of CenterPoint~\cite{yin2021center} and our BOTT. 

\begin{figure}[!h]
\centering
\footnotesize
\setlength{\tabcolsep}{0.02cm}
{\renewcommand{\arraystretch}{1}
\begin{tabular}{p{0.25cm}p{0.22\textwidth}p{0.22\textwidth}}
&\raisebox{-0.1\height}{(a)~frame $I^i$} & \raisebox{-0.1\height}{(b)~frame $I^{i+18}$}
\\
{\rotatebox[origin=c]{90}{Moving Car}}&\raisebox{-0.4\height}{\includegraphics[width=0.22\textwidth,height=0.22\textwidth,frame]{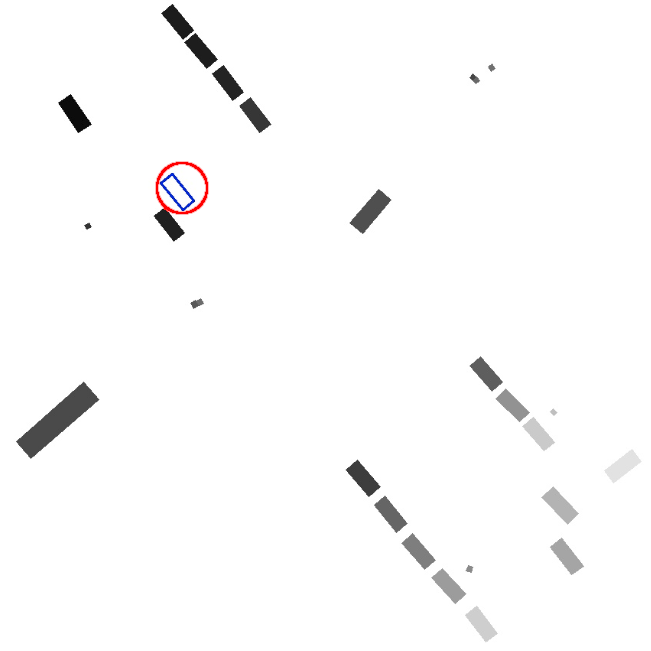}} & \raisebox{-0.4\height}{\includegraphics[width=0.22\textwidth,height=0.22\textwidth,frame] {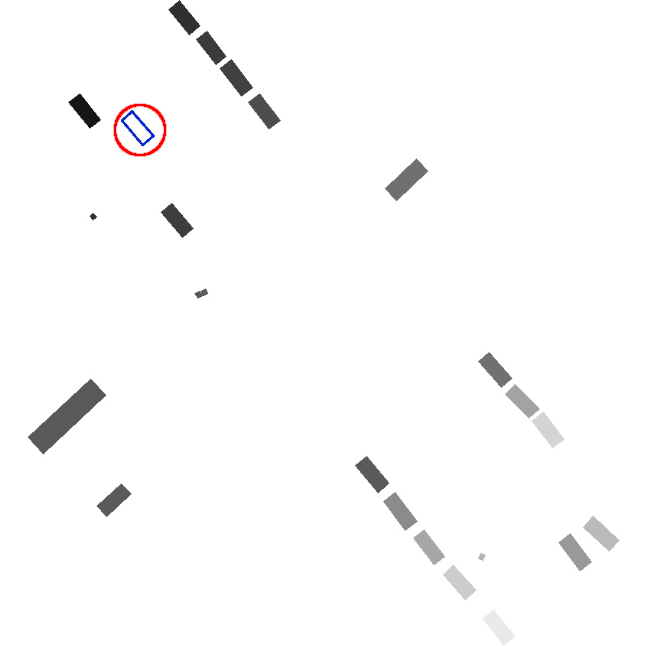}}
\\ 
&\raisebox{-0.1\height}{(c)~frame $I^j$} & \raisebox{-0.1\height}{(d)~frame $I^{j+11}$}
\\
{\rotatebox[origin=c]{90}{Static Pedestrian}}&\raisebox{-0.4\height}{\includegraphics[width=0.22\textwidth,height=0.22\textwidth,frame]{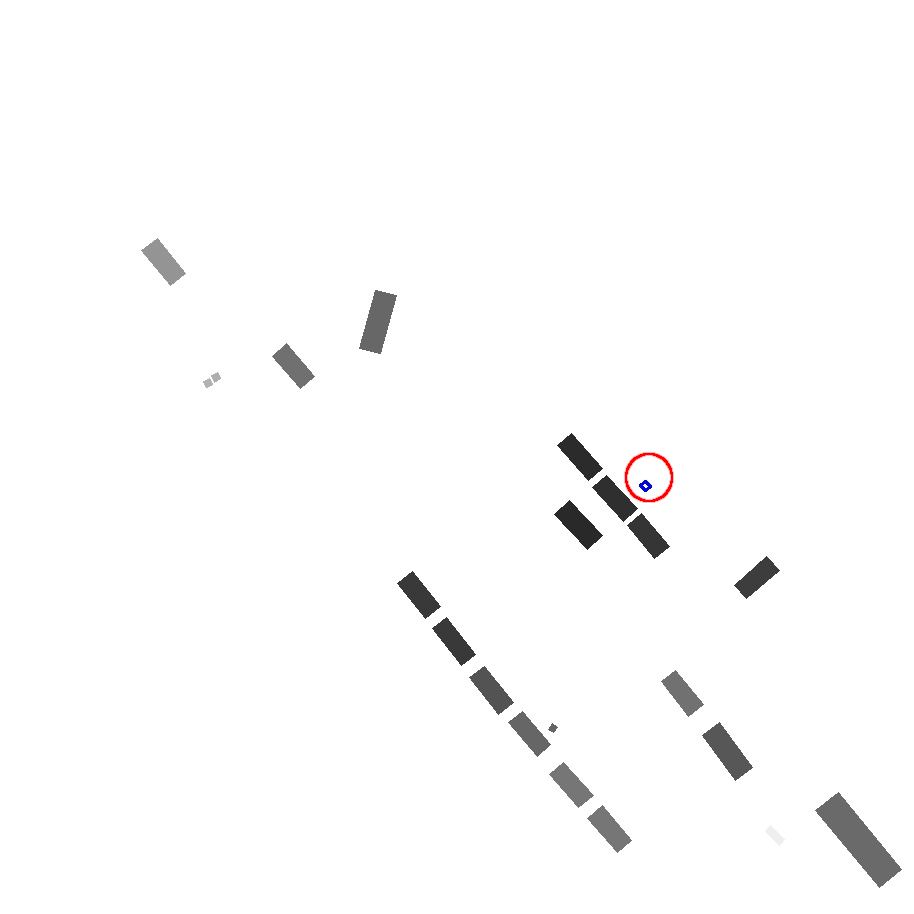}} & \raisebox{-0.4\height}{\includegraphics[width=0.22\textwidth,height=0.22\textwidth,frame]{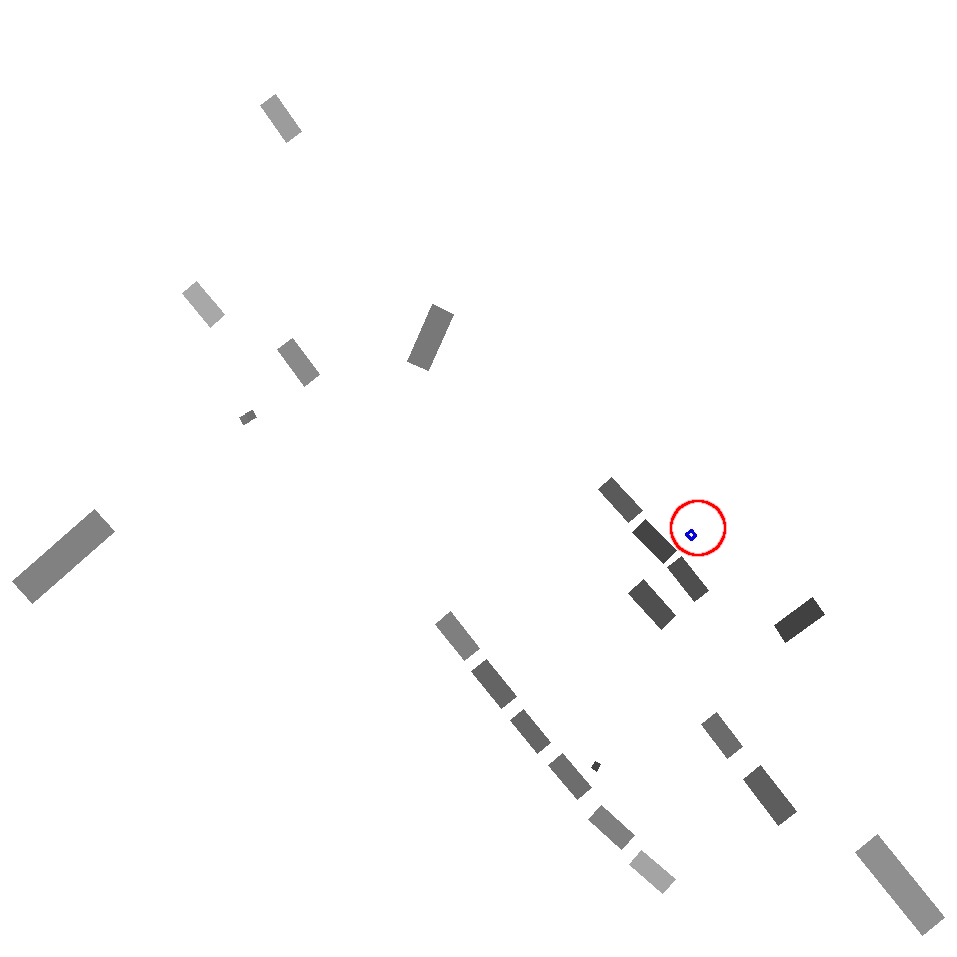}} \\
\end{tabular}}
\caption{Visualization of attention weights of the circled box. Each sub-image shows the logarithmic attention weights from other boxes, with values increasing from bright to black graysales.
}
\label{fig:attentions}
\end{figure}

Figure.~\ref{fig:attentions} showcases a few examples of the attentive boxes from the same frame that relate to the circled reference boxes. We could see that the boxes which are closer to the reference boxes normally have stronger attention impact than the boxes far away. Nevertheless, faraway boxes also contributes to the reference box due to the global self-attention. Notably, the figure only shows attentive boxes from the same frame for illustration convenience, in fact all boxes in the sliding window contribute to the reference box for a robust box embedding.

\vspace{-0.4cm}
\section{Ablation Studies}
\label{sec:ablation}
\vspace{-0.2cm}
We conduct ablation studies on nuScenes val set to analyse the impact of attention and physical constraints, and study the generalization capability of BOTT.
\subsection{Attention mechanism}
\vspace{-0.2cm}
We ablate the necessity of using the attention mechanism to learn the context information. Initially, we remove the attention layers and relied solely on raw box features to establish the links between all the boxes. We also conducted experiments with different encoder layers. The ablation results for the attention mechanism on the nuScenes validation set are presented in Table.~\ref{tab:ablation_attention}. The results revealed that without attention, the performance is worse than our best setting in Table.~\ref{tab:nuscenes_benchmark}, with a $1.80
$ AMOTA drop. It is observed that we can achieve best performance with 3 encoder layers, and further adding attention layers in the encoder does not improve the performance. 
\begin{table}[!htb]
\caption{Effect of encoder layers on nuScenes val split.}
\label{tab:ablation_attention}
\footnotesize
\centering
\begin{tabular}{|c|ccc|}
\hline
 \# encoders layers & AMOTA$\uparrow$ & IDS$\downarrow$ & Recall$\uparrow$ \\
\hline\hline
0 & 68.0 & 734 & 69.9 \\[-0.06cm]
1 & 68.76 & 520 & 70.4 \\[-0.06cm]
3 & \textbf{69.91} & \textbf{438} & \textbf{72.1} \\[-0.06cm]
6 & 69.76 & 455 & 70.4 \\
\hline
\end{tabular}
\vspace{-0.2cm}
\end{table}
\subsection{Physical Constraints}
During deployment, physical constraints are added to restrain some faraway boxes from linking together, including 1) maximal velocity constraint: center distance of two boxes could not exceed the product of time difference and object category maximal velocity; and 2) static position constraint, static boxes (with absolute speed less than $0.5m/s$ from CenterPoint\cite{yin2021center} detections) are prohibited to link boxes above 2 meters. Table.~\ref{tab:physical_constraints} compares the tracking results on nuScenes val set with different constraints. The results shows it could benefit tracking performance by reducing the interference from the impossible links. This agrees with the observation in PolarMOT~\cite{kim2022polarmot} that constructing sparse graph using boxes within certain range improves tracking performance.
\begin{table}[!htb]
\centering
\caption{Effect of physical constraints on nuScenes val set.}
\label{tab:physical_constraints}
\footnotesize
\begin{tabular}
{|c|ccc|}
\hline
Constraints & AMOTA$\uparrow$ & IDS~$\downarrow$  & Recall~$\uparrow$  \\
\hline\hline
none & 64.84  & 1485 & 68.3 \\ [-0.06cm]
velocity & 67.4 & 927 & 71.3  \\ [-0.06cm]
position & 67.1 & 1245 & 71.7  \\ [-0.06cm]
pos. + vel. & \textbf{69.91} & \textbf{438} & \textbf{72.1}  \\
\hline
\end{tabular}
\vspace{-0.2cm}
\end{table}

\subsection{Generalization Studies}

\begin{table*}[!htb]
\tiny
\centering
\caption{BOTT generalization results between nuScenes and WOD datasets.}
\label{tab:ablation_generalization}
\footnotesize
\begin{tabular}{|l|c|ccc|c|ccc|}
\hline
 & \multicolumn{8}{c|}{Evaluation} \\
\cline{2-9}
Training &  \multicolumn{4}{c|}{nuScenes} & \multicolumn{4}{c|}{WOD} \\
\cline{2-9}
& class & AMOTA$\uparrow$ & IDS$\downarrow$ & Recall$\uparrow$ & class & MOTA(L1)$\uparrow$ & MOTA(L2)$\uparrow$ & Mismatch (L2)$\downarrow$  \\
\hline\hline
nuScenes & \multirow{2}{*}{car} & 83.1 & \textbf{181}  & 82.2 & \multirow{2}{*}{vehicle} & 58.2 & 54.68  & 0.53 \\
WOD &  & \textbf{83.2} & 222 & \textbf{84.5} &  & \textbf{58.62}  & \textbf{55.08}  & \textbf{0.14}  \\
\hline
nuScenes & \multirow{2}{*}{ped} & \textbf{78.2} & 360  & \textbf{82.6} & \multirow{2}{*}{ped} & 59.83 & 55.92  & 1.23 \\
WOD &  & 77.2 & \textbf{308} & 76.3 &  & \textbf{60.35} & \textbf{56.42}  & \textbf{0.73} \\
\hline
nuScenes & \multirow{2}{*}{bicycle} & \textbf{49.8} & 1  & 47.7 & \multirow{2}{*}{cyclist} & 57.02 & 56.95  & 0.67 \\
WOD &  & 48.1 & 1 & \textbf{48.3} &  & \textbf{57.55} & \textbf{57.49}  & \textbf{0.38} \\
\hline
\end{tabular}
\vspace{-0.5cm}
\end{table*}

 \noindent \textbf{Cross-dataset Generalization}: This study shows how the BOTT online tracker generalizes across datasets. The nuScenes and WOD benchmarks have different input frequencies and number of classes, and slightly distinct annotation instructions. For example, vehicle in WOD includes any object that can be recognized as a vehicle, including motorcyles; Also parked bicycles are not labelled in WOD but they are labelled in nuScenes. To mitigate it, we conducted following processing steps: 1) nuScenes model was trained with 3 classes by merging vehicle categories: (car, bus, trailer, truck, motorcycle)$\rightarrow$ cyclist; 2) both nuScenes and WOD use 10Hz for training and validation; and 3) only evaluate car, pedestrian and bicycle in 7-class nuScenes validation set as it is not able to evaluate the merged vehicle category in nuScenes benchmark.  Table.~\ref{tab:ablation_generalization} reports the class-specific generalization results of BOTT online tracker between nuScenes and WOD. BOTT shows very promising generalization capability between nuScenes and WOD for all categories: the model trained with source dataset has slightly inferior performance than the model trained with target dataset.
\begin{table}[!htb]
\centering
\caption{Effect of input frequencies on nuScenes val.}
\label{tab:ablation_db_freq}
\footnotesize
\begin{tabular}{|c|c|ccc|}
\hline
 Frequency & K & AMOTA$\uparrow$ & IDS$\downarrow$ & Recall$\uparrow$ \\
\hline\hline
2Hz & 4 & 68.7 & 992 & 73.2 \\
5Hz & 8 & 68.6 & 617 & 71.5 \\
10Hz* & 16 & \textbf{69.9} & \textbf{438} & \textbf{72.1} \\
20Hz & 32 & 68.8 & 529 & 71.3 \\
\hline
\end{tabular}
\begin{tablenotes}
\item *Model training frequency
\end{tablenotes}
\vspace{-0.2cm}
\end{table}

 \noindent \textbf{Input Frequency Generalization}: This ablation is to assess whether BOTT is sensitive to input frequencies. CenterPoint~\cite{yin2021center} is deployed to generate 20Hz detections for nuScenes validation set, and then generate corresponding 20Hz track database following Section~\ref{sec:track_db_gen}. Afterwards, validation track databases at 10Hz, 5Hz and 2Hz are generated by keep every 2, 4 and 10 frames from the 20Hz database. To test model generalization against input frequency, a BOTT model was trained with 10Hz track database with sliding window size $K\mathrm{=}16$, and was tested with validation track databases at 20Hz, 10Hz, 5Hz and 2Hz, with $K$ accordingly being adjusted to 32, 16, 8 and 4. The results reported in Table.~\ref{tab:ablation_db_freq} is encouraging that the shared BOTT model achieves fairly good performance at various input frequencies and window sizes. In practice, this enables that BOTT tracker has higher deployment flexibility on input data.
\section{Conclusion}
\label{sec:conclusion}
We propose BOTT, a novel machine learning based 3D MOT method that relies only on 3D bounding boxes. BOTT leverages a transformer encoder with global self-attention to learn box features enriched with spatial-temporal context information from all the boxes across a time window, leading to a significant performance boost. Notably, BOTT does not rely on appearance information from raw data, which leads to better generalization ability. BOTT can be easily configured as an online or offline tracker, and our experiments demonstrate achieves state-of-the-art performance among the learning-based trackers on nuScenes and WOD. 
{\small
\bibliographystyle{ieee_fullname}
\bibliography{references}
}
\typeout{}
\clearpage

\renewcommand{\baselinestretch}{1}
\setlength{\belowcaptionskip}{0pt}
\begin{center}
\vspace{-5ex}
\Large{\bf BOTT: Box Only Transformer Tracker for 3D Object Tracking}
\\
\Large{\bf- Supplementary Material -}\\
\end{center}

\setcounter{section}{0}
\setcounter{equation}{0}
\setcounter{figure}{0}
\setcounter{table}{0}
\setcounter{page}{1}
\makeatletter

\renewcommand{\thesection}{A.\arabic{section}}
\renewcommand{\thesubsection}{A.\arabic{subsection}}
\renewcommand{\thetable}{A.\arabic{table}}
\renewcommand{\thefigure}{A.\arabic{figure}}

\normalsize
In this appendix, we (i) provide more details for track dataset generation; (ii) study impact of various data augmentations; (iii) analyse runtime breakdown of online tracker; and (iv) provide more details for online tracking.

\section{Track database generation details}
This section presents further details about track database generation. 
The track database is generated by matching detection boxes and ground truth provided by nuscenes and WOD. We follow 4 steps to generate the track database. Firstly, we deploy CenterPoint~\cite{yin2021center} on the training, validation and test sets on nuScenes and WOD following the official instruction at 
\url{https://github.com/tianweiy/CenterPoint}. The pretrained model VoxelNet with flip augmentation
is adopted to deploy on the nuScenes.
As we generated the track database at 10Hz, we downsample the input point clouds, but keep all the key frames. Due to nuScenes only provides 2Hz GT, we interpolate the GT to 10Hz. As for WOD, we use the two-stage VoxelNet model with velocity.
Secondly, as the raw detection boxes have many overlapped boxes and low-score objects, we perform both NMS and filtering of low-score objects to the raw detection boxes. The detailed parameters used for nuScenes and WOD are listed in Table.~\ref{tab:bottdb_gen_param}.
Then, we perform a class-aware Hungarian association between the filtered detection boxes and GT boxes. A box pair is only determined to be matched if interaction over union (IoU) between them is higher than the minimal threshold (0.0001). For the matched pair, the detection box will be assigned the track ID of the matched GT. Finally, all the detection boxes and GT boxes in a scene, the matched information between them, and other meta information are stored in an individual file, indexed by each time frame. 

\begin{table}[!htb]
\vspace{-0.2cm}
\centering
\caption{Parameters to filter raw detection boxes.}
\label{tab:bottdb_gen_param}
\footnotesize
\begin{tabular}{|l|c|c|c|c|c|c|}
\hline
\multirow{2}{*}{Param} & \multicolumn{3}{c|}{nuScenes} & \multicolumn{3}{c|}{WOD} \\
\cline{2-7}
& car & ped & other & vehicle & pedestrian & cyclist  \\
\hline\hline
NMS & 0.1 & 0.25 & 0.1 & 0.1 & 0.25 & 0.1\\ [-0.06cm]
Score & 0.2 & 0.2 & 0.1 & 0.2  & 0.2 & 0.1\\
\hline
\end{tabular}
\vspace{-0.5cm}
\end{table}

\section{Ablation Study on Data Augmentation}
This section presents ablation study on data augmentation. We conducted ablation study on three different augmentation methods including dropping tracks, global flipping, and global yaw augmentation. From the results in Table.~\ref{tab:data_augmentation}, it boosts the performance by dropping some tracks randomly. Global flipping along both x-axis and y-axis could improve the performance. To have larger diversity, global flipping on x-axis and y-axis are enabled. Global rotation is applied with yaw angle uniformly drawn between 0 and a maximal value. As seen from the results, all experiments, i.e. rotation with a maximum of 30, 60, 90, and 180 degrees, could improve the performance. Since the results are close, we just choose 90 degrees for larger diversity. We do not enable global jittering, due to jittering on all boxes will not change the box raw feature encoding with $xy \mathrm{-} xy_{min}$ as described in Section.~\ref{sec:box_encod}.

\begin{table}[!htb]
\vspace{-0.2cm}
\centering
\caption{Effect of data augmentation on nuScenes val set.}
\label{tab:data_augmentation}
\footnotesize
\begin{tabular}
{|c|ccc|}
\hline
Augmentation & AMOTA$\uparrow$ & IDS~$\downarrow$  & Recall~$\uparrow$  \\
\hline\hline
none & 68.96  & 509 & 70.9\\ [-0.06cm]
drop track & 69.38 & 434 & 71.8  \\ [-0.06cm]
\hline
xflip & 69.79 & 537 & 72.7  \\ [-0.06cm]
yflip & 69.65 & 479 & 73.3  \\ [-0.06cm]
xyflip & 69.40 & 572 & 72.0  \\ [-0.06cm]
\hline
yaw30 & 69.67 & 522 & 71.5  \\ [-0.06cm]
yaw60 & 69.07 & 507 & 71.5  \\ [-0.06cm]
yaw90 & 69.56 & 502 & 72.2  \\ [-0.06cm]
yaw180 & 69.85 & 505 & 71.6  \\ [-0.06cm]
\hline
drop+xyflip+yaw90 & 69.91 & 438 & 72.1  \\ 
\hline
\end{tabular}
\vspace{-0.5cm}
\end{table}

\section{Runtime of BOTT Online Tracking}
Realtime tracking is critical for autonomous driving. Table.~\ref{tab:bott_runtime} reports detailed runtime for BOTT online tracker on a desktop equipped with a 3.6GHz CPU and a GTX 1080-Ti GPU. Overall, online tracking runtime mainly consists of data loading time, network inference time, and box tracking time. It shows the BOTT tracker can run fast. For instance, BOTT network with 3 encoder layers can track at a fps of 47.8Hz on nuScenes val set with an average of 802 boxes per frame. In comparison, when running the official AB3DMOT~[21] and SimpleTrack~[14] code on nuScenes validation set, they require 322.1ms and 1128.4ms per frame, respectively. The slower runtime is due to (1) class-wise tracking for 7 classes, resulting in the summation of processing time (PolarMOT also trains different networks for different classes); 
(2) The Generalized IoU between boxes is computationally expensive. Additionally, SimpleTrack employs a two-stage association approach to boost performance with runtime sacrifice. In contrast, BOTT uses a single network for class-agnostic affinity matrix computation. 
\begin{table}[!htb]
\vspace{-0.2cm}
\centering
\caption{Runtime breakdown of BOTT online tracking.}
\label{tab:bott_runtime}
\footnotesize
\begin{tabular}{|c|c|c|c|c|}
\hline
$\#$encoder & Data (ms) & Network (ms) & Tracking (ms) & fps \\
\hline\hline
0 & \multirow{4}{*}{1.8} & 2.6 & 9.5 & 71.9 \\ [-0.06cm]
1 &  & 4.7 &  9.4 & 62.9  \\ [-0.06cm]
3 &  & 8.6 & 10.5 & 47.8  \\ [-0.06cm]
6 &  & 13.6 & 10.0 & 39.4 \\
\hline
\end{tabular}
\vspace{-0.3cm}
\end{table}

Table~\ref{tab:runtime_box_num} shows BOTT inference time with different box numbers. In real-world driving scenarios, it is uncommon to encounter more than 3000 boxes. Various optimizations, such as half-precision and TensorRT inference, can be applied to further reduce the time cost.
\begin{table}[!htb]
\vspace{-0.4cm}
\centering
\caption{\footnotesize BOTT inference time with various numbers of input box.}
\label{tab:runtime_box_num}
\footnotesize
\begin{tabular}{|c|ccccc|}
\hline
  box number   & 500 & 1000 & 2000 & 3000 & 4000 \\
\hline\hline
model time (ms) & 3.32 & 7.84  & 23.43 & 53.94 & 85.44 \\ 
\hline
\end{tabular}
\vspace{-0.4cm}
\end{table}
\section{Additional Details for Online Tracking}
This section provides more details for the BOTT online tracking described in~\ref{sec:online_tracker}. As mentioned, velocity physical constraints are applied to forbid faraway box links. Specifically, box center distance cannot exceed the product of class-specific maximal speed and the time difference. The maximal speed for each class are: bicycle=20$m/s$, pedestrian=10$m/s$, car=bus=motorcycle=trailer=truck=35$m/s$. Meanwhile, all links with center distance below following thresholds will be used for data association: bicycle=2m, pedestrian=1.5m, car=bus=motorcycle=trailer=truck=3m. During detection-track association, associated pairs with cost $AS^t_{ij}$ below thresholds  are used as matched pairs. Correspondingly, the minimal linking scores for each class are: bicycle=0.6, car=0.4, pedestrian=car=bus=motorcycle=trailer=truck=0.5.
\section{Sliding Window Size}
Table~\ref{tab:ablation_wnd_size} shows how different window sizes $K$ affect results. Two models are evaluated. The left side is trained with the same window size as the deployment ($K_t\mathcal{=}K_d$). The right side was trained with $K_t\mathcal{=}16$ but deployed with different $K_d$. Smaller window sizes result in less temporal context accessibility, while larger window sizes require the model to have a larger capacity to learn affinity between input boxes that grow quadratically, leading to increasing computations.
\begin{table}[!htb]
\footnotesize
\vspace{-0.2cm}
\centering
\caption{\footnotesize Effect of $K_t$ and $K_d$ on nuScenes validation set.}
\label{tab:ablation_wnd_size}
\footnotesize
\begin{tabular}{|c|c|p{0.9cm}p{0.6cm}||c|c|p{0.9cm}p{0.6cm}|}
\hline
$K_t$ & $K_d$ & AMOTA$\uparrow$ & IDS$\downarrow$ & $K_t$ & $K_d$ & AMOTA$\uparrow$ & IDS$\downarrow$ \\
\hline\hline
4 & 4 & 66.4  & 1369  & 16 & 4 & 65.8 & 1586 \\
8 & 8 & 68.8  & 818  & 16 & 8 & 67.9 & 889 \\
16 & 16 & \textbf{69.9} & \textbf{438} & 16 & 16 & \textbf{69.9} & \textbf{438}  \\
24 & 24 & 69.8 & 483 & 16 & 24 & 68.1 & 663 \\
\hline
\end{tabular}
\vspace{-0.6cm}
\end{table}

\end{document}